\documentclass{article}

\usepackage[a4paper, total={5in, 8in}]{geometry}
\usepackage[english]{babel}
\usepackage[utf8]{inputenc}
\usepackage[square,numbers]{natbib}
\usepackage{subcaption}
\usepackage{graphicx}
\usepackage{authblk}
\usepackage{xcolor}
\usepackage{hyperref}
\usepackage{amsmath,amssymb}
\usepackage{listings}
\usepackage{bibentry}
\usepackage{soul}
\usepackage[mathlines]{lineno}

\DeclareMathOperator{\E}{\mathbb{E}}

\nobibliography*
\graphicspath{{figure/}}

\definecolor{code_bg}{HTML}{f5f5f5}
\newcommand{\incode}[1]{%
  \lstset{basicstyle=\ttfamily,breaklines=true,columns=fullflexible}%
  \lstinline|#1|}

\title{TensorFlow Agents: Efficient Batched Reinforcement Learning in TensorFlow}
\author{Danijar Hafner\footnote{Work done during an internship with Google Brain.}, James Davidson, Vincent Vanhoucke\\Google Brain\\\texttt{mail@danijar.com, jcdavidson@google.com, vanhoucke@google.com}}
\date{August 24, 2017}

\begin{document}

\maketitle

\begin{abstract}
We introduce TensorFlow Agents, an efficient infrastructure paradigm for building parallel reinforcement learning algorithms in TensorFlow. We simulate multiple environments in parallel, and group them to perform the neural network computation on a batch rather than individual observations. This allows the TensorFlow execution engine to parallelize computation, without the need for manual synchronization. Environments are stepped in separate Python processes to progress them in parallel without interference of the global interpreter lock. As part of this project, we introduce BatchPPO, an efficient implementation of the proximal policy optimization algorithm. By open sourcing TensorFlow Agents, we hope to provide a flexible starting point for future projects that accelerates future research in the field.
\end{abstract}

\section{Introduction}
\label{introduction}

We introduce a unified interface for reinforcement learning agents with accompanying infrastructure integrating with TensorFlow~\cite{abadi2016tensorflow} that allows to efficiently develop new algorithms. Defining a standard algorithm interface for reinforcement learning allows us to reuse common infrastructure between all algorithms, and to change the algorithm as easily as the environment. We aim to accelerate future reinforcement learning research by releasing this project to the public.\footnote{\url{https://github.com/brain-research/batch-ppo}}

The most important goals of this project are to make it easy to implement new algorithms and to train them fast. Many reinforcement learning methods spend most of their time interacting with the environment, as compared to the time required to compute and apply gradient updates. There are two bottlenecks during the simulation of the algorithm and the environment: The forward pass of the neural network and progressing the environment. We address both limitations by simulating many environments in parallel, and by combining their observations for the algorithm to operate on batches of data.

To simulate multiple environments in parallel, we extend the OpenAI Gym interface~\cite{brockman2016gym} to batched environments that are stepped in parallel. We use separate processes for every environment, so that environments implemented in Python can run freely without being limited by the global interpreter lock. By integrating the batched environment into the TensorFlow graph, we can combine it with the reinforcement learning algorithm leaving just a single operation to call within the training loop.

We release the TensorFlow Agents project to the open source community together with BatchPPO, an optimized implementation of the recently introduced proximal policy gradient algorithm~\cite{schulman2017ppo}, a simple and powerful reinforcement learning baseline that showed impressive results in locomotion tasks~\cite{heess2017parkour}. We believe that releasing this project can speed up the progress of reinforcement learning research.

\section{Related Work}
\label{related-work}

We present a unified agent interface together with and efficient TensorFlow framework for parallel reinforcement learning and an implementation of the recently introduced proximal policy gradient algorithm~\cite{schulman2017ppo}.

With OpenAI Gym, \citet{brockman2016gym} introduced an standardized interface for environments that has since then gained wide adoption. We extend this work by providing an interface that allows to combine multiple Gym environments and step them together. Moreover, we introduce a common interface for learning algorithms implemented in TensorFlow. Our aim is that this will allow to make both environments and algorithms easily exchangeable part in a research project.

\citet{duan2016rllab} released the Rllab framework that implements several reinforcement learning algorithms and provides a common infrastructure for training and evaluation. Numerical computations are mainly implemented in Numpy and Theano. Our work also provides tools for training and evaluation, but is optimized for parallel simulation, resulting in significantly accelerated training.

OpenAI Baselines \citep{sidor2017baselines} aims to provide high-quality implementations of reinforcement learning algorithms. Among other algorithms, it includes an implementation of PPO using TensorFlow for the neural network computation. This implementation relies on Python for most of the algorithm logic which is accelerated using Mpi4py~\cite{dalcin2017mpi}. TensorFlow Agents has a similar goal, but provides reusable infrastructure for future vectorized implementations of reinforcement learning algorithms.

\citet{schulman2017modularrl} provides an implementation of PPO written in Numpy and Keras. In addition to the gradient-based variant described in this paper, it includes a version of PPO that uses LBFGS for updating the neural networks. This project served as a reference for the algorithmic part of our BatchPPO implementation. We build upon this work by parallelizing the experience collection and providing TensorFlow infrastructure that is reusable for further research projects.
\section{Background}
\label{background}

In this section we summarize the background required to understand the proximal optimization optimization algorithm~\cite{schulman2017ppo}, of which we provide an efficient implementation that we call BatchPPO.

We consider the standard reinforcement learning setting defined as a Partially Observable Markov Decision Process $POMDP=\langle\,S,\Omega,A,P,R,\mathcal{O},\gamma,s_0\,\rangle$ that defines a stochastic sequence of states $s_t\in S$, observations $x_t\in\Omega$, actions $a_t\in A$, and rewards $r_t\in\mathbb{R}$. Starting from the initial state $s_0$, we draw observations $x_t\sim\mathcal{O}(s_t)$, actions $a_t\sim\pi(a_t|x_{1:t})$ from the policy, rewards $r_t\sim R(r_t|s_t,a_t)$ from the reward function, and the following state $s_{t+1}\sim P(s_{t+1}|s_t,a_t)$ from the transition function. The objective is to find a policy $\pi$ to maximize the expectation of the return $V^{\pi}(s_t)=\E_{\pi}\big[\sum_{i=0}^{\infty}{\gamma^i r_{t+i}}\big]$.

We mainly consider the case where $S$ and $A$ are continuous. In this case, it is natural to parameterize the policy by a neural network and learn its weights to maximize the empirical objective $R_t=\sum_{i=0}^{\infty}{\gamma^i r_{t+i}}$ at each time step. The objective for training the neural network is given by \citet{williams1992reinforce} as:

\begin{equation}
L^{PG}(\theta)=\E_{\pi_{\theta}}{\big[\ln \pi_{\theta}(a_t|x_{0:t})R_t\big]}.
\end{equation}

We can estimate the expectation by collecting multiple episodes from the environment. To use the collected episodes for multiple gradient steps, we need to correct for the changing action distribution of the updated policy $\pi_{\theta}$ from the data collecting policy $\pi'$ using importance sampling:

\begin{equation}
L^{IS}(\theta)=\E_{\pi'}{\bigg[\frac{\pi_{\theta}(a_t|x_{0:t})}{\pi'(a_t|x_{0:t})}R_t\bigg]}.
\end{equation}

Due to stochasticity in the transition function, reward function, and our policy, we would need to collect many episodes for a sufficient estimate of this expectation. We can reduce the variance of our estimate without introducing bias by subtracting the state values $V^{\pi'}(s_t)=\E_{\pi'}[\sum_{i=0}^{\infty}{\gamma^i r_{t+i}}]$ from the returns $R_t$. This effectively removes the stochastic influences of the following time steps from the estimate:

\begin{equation}
L^{VB}(\theta)=\E_{\pi'}{\bigg[\frac{\pi_{\theta}(a_t|x_{0:t})}{\pi'(a_t|x_{0:t})}\big(R_t-V^{\pi'}_{\theta}(s_t)\big)\bigg]}.
\end{equation}

Finally, we need to prevent updates from changing the action distribution too much at once. To accomplish this, PPO penalizes the KL change between the policy before the update $\pi'$, and the policy at the current update step $\pi_{\theta}$. The penalty weight $\beta$ is adjusted based on the observed change in KL divergence after performing all updates on the collected batch of episodes. The final objective reads:

\begin{equation}
L^{KL}(\theta)=\E_{\pi'}{\bigg[\frac{\pi_{\theta}(a_t|x_{0:t})}{\pi'(a_t|x_{0:t})}\big(R_t-V^{\pi'}(s_t)\big)\bigg]}+\beta\E_{\pi'}{\Big[D_{KL}\big(\pi'(a_t|x_{0:t})||\pi_{\theta}(a_t|x_{0:t})\big)\Big]}.
\end{equation}

In practice, we collect a small number of episodes, perform several updates of full-batch gradient descent on it, and then discard the collected data and repeat this process. \citet{schulman2017ppo} also proposed an alternative objective that clips the importance sampling ratio, but we found the objective using the KL penalty to work very well in practice and did not explore this option.
\section{Parallelism}
\label{parallelism}

In many cases, the bottleneck of reinforcement learning algorithms is collecting episodes from the environment. There are two time consuming computations in this process: The forward pass of the neural network and stepping the environments. Fortunately, we can parallelize both processes by using multiple environments at the same time. Previous implementations use asynchronous workers for this~\cite{mnih2016a3c}, but we can fully utilize the system with a simpler architecture while saving communication overhead. Our implementation is completely defined within the TensorFlow graph.

First, we parallelize the forward pass of the neural network by vectorizing the agent computation to produce a batch of actions from a batch of observations. This is similar to Facebook's ELF~\cite{tian2017elf}, where agents also process multiple observations at each step. Using a batch size during inference allows us to leverage the internal thread pool of the TensorFlow session, or hardware accelerators such as GPUs and TPUs.

Second, we simulate environments in parallel. Relying on the TensorFlow session is not sufficient for this purpose, since many environments are implemented in Python and are restricted by the Global Interpreter Lock (GIL). Instead, we spawn a separate process for each environment, so that all available CPU cores can step environments in parallel. Communication between the main process and the environment processes introduces a constant overhead, but this is alleviated by the ability to run many environments in parallel.

While we parallelize the forward passes of the neural network and the environment stepping, we do not allow environments to step faster than others and get out of sync. This makes algorithm implementations conceptually simple yet efficient. When training on CPUs, this implementation fully utilizes the available resources. When using GPUs, the system switches between full load for the CPUs and the GPUs. This process could potentially be parallelized by introducing an action lag so that the environments can step together with the algorithm.
\section{Implementation}
\label{implementation}

We introduce an interface and infrastructure for algorithms for efficiently interacting with parallel environments. We first describe the algorithm interface, followed by the batched environment interface, and the simulation operation combining the two.

\subsection{Algorithm}

A TensorFlow Agents algorithm defines the inference and learning computation for of a batch of agents. It implements the following functions that define the computation graph of the algorithm:

\begin{itemize}
\item\incode{begin_episodes(agent_indices)} is called with a list of the agents that start a new episode at the current time step. The agents are represented as indices into the batch dimension.
\item\incode{perform(observation)} defines the batched inference computation of the algorithm, and returns an actions as a tensor with batch dimension.
\item\incode{experience(observation, action, reward, done, next_obs)} lets the algorithm process the batch of transition tuples after the environments have been stepped.
\item\incode{end_episodes(agent_indices)} is called for each batch index that finished an episode at the current time step.
\end{itemize}

These operations additionally return string tensors containing TensorFlow summaries. Our infrastructure combines these summaries and writes them jointly. If these tensors are empty strings, no summaries will be written at the current step.

We provide the BatchPPO implementation defined in \incode{ppo.PPOAlgorithm(envs, step, is_training, should_log, config)} as an example implementation of this interface. The constructor expects an in-graph batch environment, a tensor of the global step, a tensor flag indicating whether to compute summaries at the current time step, and a configuration object.

The neural networks used by BatchPPO are implemented as \incode{tf.contrib.rnn.RNNCell()} classes, and are specified in the configuration object. Custom neural network structures can be defined by implementing the \incode{tf.contrib.rnn.RNNCell()} interface, returning a tuple containing tensors for the action mean, log action standard deviation, and state value. We chose this interface as it integrates nicely into the existing TensorFlow ecosystem, while providing freedom in the model optimized by the algorithm. We provide example implementations for a feed forward policy and a recurrent policy.

\subsection{Environments}

To efficiently simulate environments in parallel, we provide the \incode{agents.tools.wrappers.ExternalProcess(constructor)} environment wrapper. This wrapper constructs an OpenAI Gym environment inside of an external process. Calls to \incode{step()} and \incode{reset()}, as well as attribute accesses, are forwarded to the environment and wait for the result. This environment wrapper is compatible with all existing Python environments that comply to the OpenAI Gym interface~\cite{brockman2016gym}, as long as they do not rely on shared global state.

The \incode{agents.tools.BatchEnv(envs, blocking)} class extends the OpenAI Gym interface to vectorized environments. It combines multiple OpenAI Gym environments, with \incode{step()} accepting a batch of actions and returning batches of observations, rewards, done flags, and info objects. If the individual environments live in external processes, they will be stepped in parallel. The observation and action spaces of the combined environments must all match and can be accessed without modification. In addition, the batch environment allows to access individual environments via index. Alternatively, custom environments can be implemented in vectorized format directly.

We integrate both the environment and algorithm into the TensorFlow graph. This allows us to prevent copying data and to perform simulation with a small number of TensorFlow session runs. The class \incode{agents.tools.InGraphBatchEnv(batch_env)} integrates a batch environment into the TensorFlow graph and makes its \incode{step()} and \incode{reset()} functions accessible as operations. The current batch of observations, last actions, rewards, and done flags is stored in variables and made available as tensors. All observations and rewards are converted to 32-bit data types, as these are expected by most neural network implementations.

\subsection{Simulation}

The TensorFlow operation \incode{tools.simulate(envs, algo, log, reset)} fuses together an in-graph batch environment and an algorithm. The optional \incode{log} and \incode{reset} tensors indicate whether the algorithm should compute TensorFlow summaries at the current step, and whether all environments should be reset so that all agent indices start new episodes. This is used when the training protocol switches between training and evaluation phases to discontinue the ongoing episodes. This operation provides a way to run a single TensorFlow session call inside the training loop, so that the training loop resembles the case of supervised learning.

\section{Experiments}
\label{experiments}

\begin{figure}
\centering
\begin{subfigure}[t]{.31\textwidth}
\centering
\includegraphics[width=\textwidth]{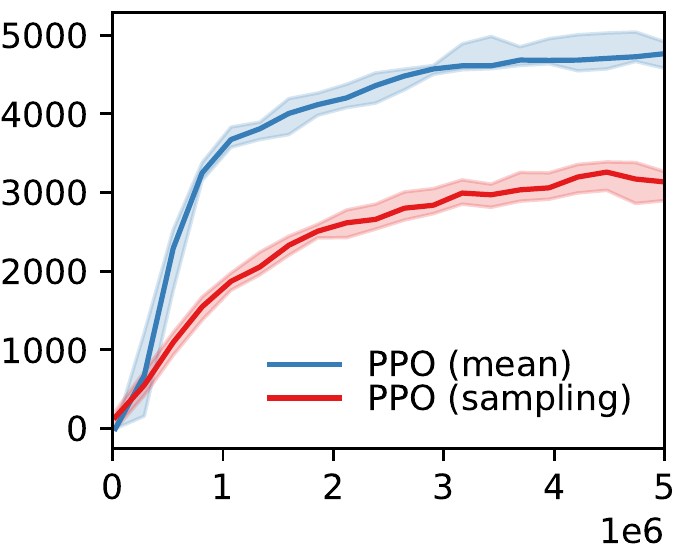}
\caption{HalfCheetah-v1}
\label{fig:cheetah}
\end{subfigure}\hfill%
\begin{subfigure}[t]{.31\textwidth}
\centering
\includegraphics[width=\textwidth]{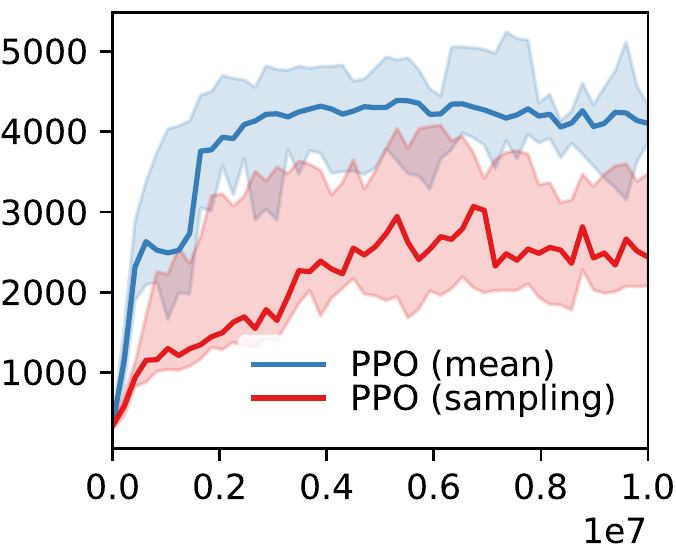}
\caption{Walker2d-v1}
\label{fig:walker}
\end{subfigure}\hfill%
\begin{subfigure}[t]{.31\textwidth}
\centering
\includegraphics[width=\textwidth]{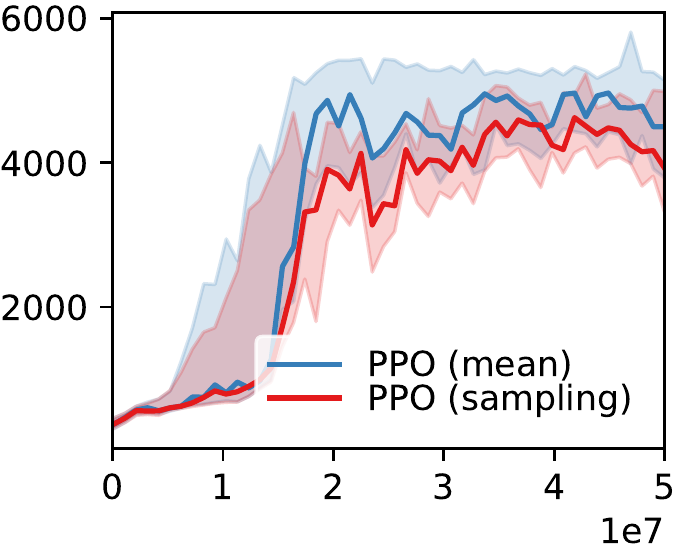}
\caption{Humanoid-v1}
\label{fig:humanoid}
\end{subfigure}%
\caption{BatchPPO episode returns by environment steps. The blue line indicates performance when using the mean action, while the red line indicated performance when sampling from the action distribution. Our results are on par or better than published results using the PPO algorithm~\cite{schulman2017ppo}.}
\label{fig:scores}
\end{figure}

To demonstrate the performance of our infrastructure and implementation, we train the BatchPPO algorithm on control tasks in the MuJoCo~\cite{todorov2012mujoco} domain. Our agent uses two neural networks, one to compute a mean action from the current observation, and one to provide an estimate of the state value. We also experimented with recurrent neural networks but generally observed slower learning and similar final performance.

The log standard deviation of the action elements is learned as an independent parameter vector. The actions are then sampled from Gaussian distribution that is parameterized by the predicted mean, and the log standard deviation used to define the diagonal elements of the covariance matrix. The elements that constitute an action vector are therefore independent. During evaluation, we use the policy's mean action, rather than sampling from the distribution.

We used the same hyper parameter configuration for the considered tasks. Specifically, we collect batches of 25 episodes before each update, and perform 25 gradient steps each for the policy and value networks using Adam optimizers~\cite{kingma2014adam} with fixed learning rates of $10^{-4}$ and $10^{-3}$ respectively. Both networks use two layers of 200 and 100 units with ReLU non-linearities~\cite{hahnloser2000relu}. We apply tanh to the action mean to fix its range. Following \citet{schulman2017ppo} and \citet{heess2017parkour}, we use streaming statistics of the observations and rewards to normalize them. We further adopt their additional cut-off penalty that we enable when overshooting the desired change in KL divergence by a factor of 2.

\begin{figure}
\centering
\begin{subfigure}[b]{.31\textwidth}
\centering
\includegraphics[width=\textwidth]{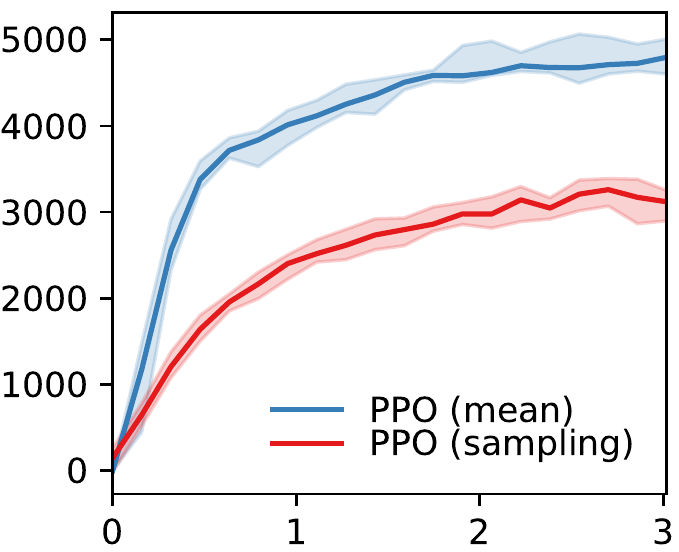}
\caption{HalfCheetah-v1}
\label{fig:cheetah-time}
\end{subfigure}\hfill%
\begin{subfigure}[b]{.31\textwidth}
\centering
\includegraphics[width=\textwidth]{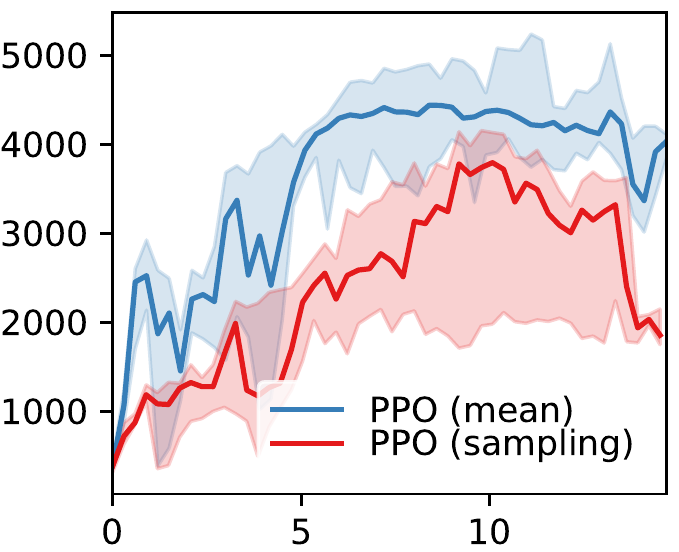}
\caption{Walker2d-v1}
\label{fig:walker-time}
\end{subfigure}\hfill%
\begin{subfigure}[b]{.31\textwidth}
\centering
\includegraphics[width=\textwidth]{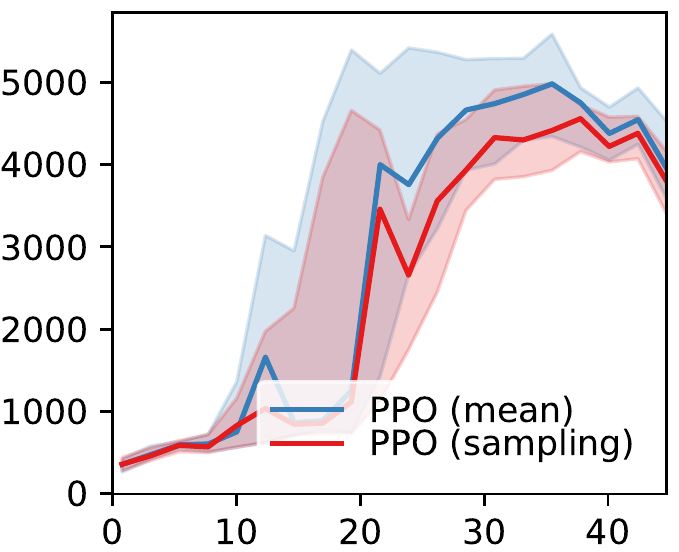}
\caption{Humanoid-v1}
\label{fig:humanoid-time}
\end{subfigure}%
\caption{BatchPPO episode returns by training time in hours using 6 CPU cores. The blue line indicates performance when using the mean action, while the red line indicated performance when sampling from the action distribution. Our implementation can quickly solve challenging locomotion tasks on a single machine.}
\label{fig:times}
\end{figure}

The results of our BatchPPO implementation are shown in Figures~\ref{fig:scores}~and~\ref{fig:times} by environment steps and hours of training time using 6 CPU cores, respectively. The blue line shows evaluation performance when acting using the mean action, while the red line shows the training performance when sampling from the policy's action distribution. Shaded areas indicate the 25\,th to 75\,th percentile over three seeds. We observe reliably high performance that is on par or better than published results from existing PPO implementations~\cite{schulman2017ppo,heess2017parkour}.

\section{Discussion}
\label{discussion}

We introduced an infrastructure paradigm and implementation using TensorFlow for parallel reinforcement learning algorithms, including BatchPPO, an efficient implementation of the proximal policy optimization algorithm. We lay out our reasoning for the design choices that result in a simple and extendable yet performant implementation. We hope that providing our infrastructure to the public can accelerate further research in reinforcement learning, and provides a powerful framework for new algorithm implementations. In the future, custom environments could be implemented in a vectorized way, possibly within TensorFlow, to leverage parallel hardware without introducing inter-process communication overhead.
\section{Acknowledgements}
\label{acknowledgements}

We thank Nicolas Heess and Josh Merel from DeepMind for insightful discussions. Furthermore, we thank the TensorFlow team and community for developing TensorFlow and thus making this project possible.

\bibliography{references}

\begin{thebibliography}{14}
\providecommand{\natexlab}[1]{#1}
\providecommand{\url}[1]{\texttt{#1}}
\expandafter\ifx\csname urlstyle\endcsname\relax
  \providecommand{\doi}[1]{doi: #1}\else
  \providecommand{\doi}{doi: \begingroup \urlstyle{rm}\Url}\fi

\bibitem[Abadi et~al.(2016)Abadi, Agarwal, Barham, Brevdo, Chen, Citro,
  Corrado, Davis, Dean, Devin, et~al.]{abadi2016tensorflow}
M.~Abadi, A.~Agarwal, P.~Barham, E.~Brevdo, Z.~Chen, C.~Citro, G.~S. Corrado,
  A.~Davis, J.~Dean, M.~Devin, et~al.
\newblock Tensorflow: Large-scale machine learning on heterogeneous distributed
  systems.
\newblock \emph{arXiv preprint arXiv:1603.04467}, 2016.

\bibitem[Brockman et~al.(2016)Brockman, Cheung, Pettersson, Schneider,
  Schulman, Tang, and Zaremba]{brockman2016gym}
G.~Brockman, V.~Cheung, L.~Pettersson, J.~Schneider, J.~Schulman, J.~Tang, and
  W.~Zaremba.
\newblock Openai gym, 2016.

\bibitem[Dalcin(2017)]{dalcin2017mpi}
L.~Dalcin.
\newblock Mpi for python: Python bindings for mpi.
\newblock \url{https://github.com/mpi4py/mpi4py}, 2017.

\bibitem[Duan et~al.(2016)Duan, Chen, Houthooft, Schulman, and
  Abbeel]{duan2016rllab}
Y.~Duan, X.~Chen, R.~Houthooft, J.~Schulman, and P.~Abbeel.
\newblock Benchmarking deep reinforcement learning for continuous control.
\newblock In \emph{International Conference on Machine Learning}, pages
  1329--1338, 2016.

\bibitem[Hahnloser et~al.(2000)Hahnloser, Sarpeshkar, Mahowald, Douglas, and
  Seung]{hahnloser2000relu}
R.~H. Hahnloser, R.~Sarpeshkar, M.~A. Mahowald, R.~J. Douglas, and H.~S. Seung.
\newblock Digital selection and analogue amplification coexist in a
  cortex-inspired silicon circuit.
\newblock \emph{Nature}, 405\penalty0 (6789):\penalty0 947, 2000.

\bibitem[Heess et~al.(2017)Heess, Sriram, Lemmon, Merel, Wayne, Tassa, Erez,
  Wang, Eslami, Riedmiller, et~al.]{heess2017parkour}
N.~Heess, S.~Sriram, J.~Lemmon, J.~Merel, G.~Wayne, Y.~Tassa, T.~Erez, Z.~Wang,
  A.~Eslami, M.~Riedmiller, et~al.
\newblock Emergence of locomotion behaviours in rich environments.
\newblock \emph{arXiv preprint arXiv:1707.02286}, 2017.

\bibitem[Kingma and Ba(2014)]{kingma2014adam}
D.~Kingma and J.~Ba.
\newblock Adam: A method for stochastic optimization.
\newblock \emph{arXiv preprint arXiv:1412.6980}, 2014.

\bibitem[Mnih et~al.(2016)Mnih, Badia, Mirza, Graves, Lillicrap, Harley,
  Silver, and Kavukcuoglu]{mnih2016a3c}
V.~Mnih, A.~P. Badia, M.~Mirza, A.~Graves, T.~Lillicrap, T.~Harley, D.~Silver,
  and K.~Kavukcuoglu.
\newblock Asynchronous methods for deep reinforcement learning.
\newblock In \emph{International Conference on Machine Learning}, pages
  1928--1937, 2016.

\bibitem[Schulman(2017)]{schulman2017modularrl}
J.~Schulman.
\newblock Modular rl: Implementation of trpo and related algorithms.
\newblock \url{https://github.com/joshu/modular_rl}, 2017.

\bibitem[Schulman et~al.(2017)Schulman, Wolski, Dhariwal, Radford, and
  Klimov]{schulman2017ppo}
J.~Schulman, F.~Wolski, P.~Dhariwal, A.~Radford, and O.~Klimov.
\newblock Proximal policy optimization algorithms.
\newblock \emph{arXiv preprint arXiv:1707.06347}, 2017.

\bibitem[Sidor et~al.(2017)Sidor, Schulman, Plappert, and
  contributors]{sidor2017baselines}
S.~Sidor, J.~Schulman, M.~Plappert, and contributors.
\newblock Openai baselines: High-quality implementations of reinforcement
  learning algorithms.
\newblock \url{https://github.com/openai/baselines}, 2017.

\bibitem[Tian et~al.(2017)Tian, Gong, Shang, Wu, and Zitnick]{tian2017elf}
Y.~Tian, Q.~Gong, W.~Shang, Y.~Wu, and L.~Zitnick.
\newblock Elf: An extensive, lightweight and flexible research platform for
  real-time strategy games.
\newblock \emph{arXiv preprint arXiv:1707.01067}, 2017.

\bibitem[Todorov et~al.(2012)Todorov, Erez, and Tassa]{todorov2012mujoco}
E.~Todorov, T.~Erez, and Y.~Tassa.
\newblock Mujoco: A physics engine for model-based control.
\newblock In \emph{Intelligent Robots and Systems (IROS), 2012 IEEE/RSJ
  International Conference on}, pages 5026--5033. IEEE, 2012.

\bibitem[Williams(1992)]{williams1992reinforce}
R.~J. Williams.
\newblock Simple statistical gradient-following algorithms for connectionist
  reinforcement learning.
\newblock \emph{Machine learning}, 8\penalty0 (3-4):\penalty0 229--256, 1992.

\end{thebibliography}

\end{document}